# Emotion recognition techniques with rule based and machine learning approaches.

Aasma Aslam[a,*], Babar Hussain[b]
Dept of Electrical and Computer Engineering
The University of New Mexico
Albuquerque, NM 87131-0001, USA
[*]aaslam@unm.edu
[b]Intel Corporation, Rio Rancho, New Mexico, USA, 87124.

**Abstract**. Emotion recognition using digital image processing is a multifarious task because facial emotions depend on warped facial features as well as on gender, age, and culture. Furthermore, there are several factors such as varied illumination and intricate settings that increase complexity in facial emotion recognition. In this paper, we used four salient facial features, Eyebrows, Mouth opening, Mouth corners, and Forehead wrinkles to identifying emotions from normal, occluded and partially-occluded images. We have employed rule-based approach and developed new methods to extract aforementioned facial features similar to local bit patterns using novel techniques. We propose new methods to detect eye location, eyebrow contraction, and mouth corners. For eye detection, the proposed methods are Enhancement of Cr Red (ECrR) and Suppression of Cr Blue (SCrB) which results in 98% accuracy. Additionally, for eyebrow contraction detection, we propose two techniques (1) Morphological Gradient Image Intensity (MGII) and (2) Degree of Curvature Line (DCL). Additionally, we present a new method for mouth corners detection. For classification purpose, we use an individual classifier, majority voting (MV) and weighted majority voting (WMV) methods which mimic Human Emotions Sensitivity (HES). These methods are straightforward to implement, improve the accuracy of results, and work best for emotion recognition using partially occluded images. It is ascertained from the results that our method outperforms previous approaches. Overall accuracy rates are around 94%. The processing time on one image using processor core i5 is ~0.12 sec.

**\*Corresponding Author**, E-mail: aasmaaslam@yahoo.com

# 1   Introduction

The translation of human emotions from facial expressions can play an important role in human-computer interaction (HCI); therefore, development of a system capable of connecting computer and human emotions is given a great importance nowadays. Facial expressions reflect the emotions and person's feelings. Especially, on different occasions, people do not want to vocalize different words due to which facial expressions gains priority over the spoken words. Therefore the facial expression recognition has gained more charm in the field of computer vision.  Due to high deformation flexibility and diversity in facial expressions, accuracies are really difficult to attend. Emotion detection can be performed using different methods such as by Interactive Voice Response (IVR). The IVR system uses audio signal analysis technique in which voice as an input is used for the detection of emotions on the basis of voice features such as loudness, pitch, use of certain words and their stress level etc [1]. This technique is not sufficiently reliable because the voice alone is not adequate for emotion detection [1]. Emotion recognition has also been addressed by analyzing Brain Activity [2], which utilizes different sensors to detect brain actions and reactions. This is one of the best techniques for the emotion detection but the main drawback of this technique is that it involves highly sophisticated sensors to detect brain signals and needs a high level of expertise. Therefore, there is a need for a reliable approach which is easy to handle and inexpensive. Use of digital images to develop emotion recognition system is the least expensive approach which can be employed in almost every environment. Digital image processing based on emotion recognition system is applicable for different purposes. For example, people having video conversation can interact better if they use the emotion recognition system. We can use this system for lie detection [3] and pain detection [4]. A potential application could be getting feedback of a person watching a TV program, which can be beneficial for the survey of people's liking or disliking about the TV programs and can provide a technology acceptance of a TV platform for the elderly living alone [5]. Another interesting application of the proposed system is regarding security, for example, gesture recognition of a person having intentions of robbery or any other offense. In the field of artificial intelligence, it can be used to model human emotions in the robots [6]. Up to some extent, it can be valuable in the field of psychology specifically in recognition of schizophrenia's status or emotions of such kind of persons [7]. In recent literature, Afifi et al. [8] did features extraction for different applications based on hand images such as for biometric identification system but the hand based approach is not as robust as the features extracted from the facial images. In this study, the emotions are recognized automatically by analyzing salient facial areas, wrinkles, eyebrows, and lips. To the best of our knowledge, the usage of forehead wrinkles along with other facial features has not been reported for emotion detection before this study. We have also defined some new rules for better identification of emotions using ensemble classifier [9] [10] by introducing Majority Voting (MV) and Weighted Majority Voting (WMV) based on Human Emotions Sensitivity (HES). In our proposed methodology, the emotion recognition system consists of three stages which are pre-processing, extraction of the required feature and classification of emotion based on these features.

Our contributions are enlisted below:

1. We have proposed a new method which we name as Enhancement of Cr Red (ECrR) and Suppression of Cr Blue (SCrB). This method works with 98% accuracy on three standard databases RaFD [11], SFEW [12][13], and JAFFE [14].

2. We have proposed two different methods Morphological Gradient Image Intensity (MGII) and Degree of Curvature Line (DCL) for extracting eyebrows feature.

3. Improvements have been performed in two existing methods for lips corners detection and lips feature extraction.

4. The majority voting and weighted majority voting methods are tuned for achieving better results. This works better for the partially occluded images.

The organization of the paper is as follows. Section 2 includes related work, section 3 contains proposed methodology, and section 4 contains discussion about RaFD database [11], SFEW database [12][13], and JAFFE database [14]. In Section 5, the optimal parameters for the classification of MV and WMV are searched out. Finally, section 6 contains conclusion of the proposed methodology.

## 2 Related Work

Ekman et al. [15] [16] classified six emotions to be universal which are, happy, disgust, sad, angry, fear, and surprise. In past, facial expression was a research subject for psychologists only, but nowadays automatic recognition of facial expressions through digital image processing has gained attention. Lyons et al. [17] present some work on images which are related to facial expressions. After Kanade et al. [18] presented their work in 2000, and the automatic recognition of facial expression gained more attention. According to Fasel et al. [19], facial emotion recognition system consists of three major phases; face detection, feature extraction, and feature classification. There are four basic approaches for face detection; Knowledge-based, Feature invariant, Template based and Appearance-based [20].

Alvino et al. [21], have measured the difference in the emotion of healthy person and the schizophrenic patient using four basic emotions. In this technique, features are extracted by 2-D Daubechies wavelet and support vector machine (SVM) is used for classification. The results are separately calculated for the males and females which are around 77% accurate. In work by Bashyal et al. [22], facial emotions are recognized using Gabor wavelet's filter for feature extraction and learning vector quantization for classification. Here, the JAFFE database is used and recognition accuracy rates are around 87.5%. Wong et al. [23] utilized a format of tree structures with Gabor feature representations to present a facial emotional state and this tree structure is processed by proposing the local expert's organization model. An Asian emotion database is also created in this study, where the results are quite robust.

Besinger et al. [24] have presented the feature point tracking technique which is applied to the five facial image regions of 60 facial images. Three facial emotions were recognized which are happy, sad and angry, and the overall accuracy rates were 83%. Bailenson et al. [25] have presented a real-time model by using machine learning algorithms. The two emotions, amusement, and sadness are recognized by facial features and different physiological responses. Calvo et al. [26] have investigated whether emotional facial expressions can be accurately identified in peripheral vision. Beneficial mechanisms for recognition of emotions and the role of eye and mouth are highlighted in the facial emotion recognition.

In another study [27], the authors have presented a simple Profilometric technique to measure the size and function of the wrinkles. Where wrinkle size was measured in relaxed conditions and four representative parameters were considered. These parameters are; the mean Wrinkle Depth, the mean Wrinkle Area, the mean Wrinkle Volume, and the mean Wrinkle Tissue Reservoir Volume (WTRV). But this is a difficult way of measurement and cannot be implemented on digital images. A study [28] has presented wrinkles detection and classification of emotions of the students based on the forehead wrinkles. Here, only the presence of wrinkles is noticed. The reported methodology is dependent on the original neutral image. In a survey [29], the authors have presented the feature extraction and discussed all the opportunities for the researchers in the facial field as well as all the challenges. In another report [30], facial features are extracted by using Prewitt edge operator, in which eyes are extracted but this method does not show promising results because it cannot handle different illumination conditions and does not work if applied directly on an image without any pre-processing steps. Different efforts have recently emerged to develop algorithms that can process naturally occurring human affective behavior [31]. Moreover, an increasing number of efforts are reported toward multimodal fusion for human affect analysis including audiovisual fusion, linguistic and paralinguistic fusion, and multi-cue visual fusion based on facial expressions, head movements, and body gestures [31].

Different classifiers are used for the classification of emotions where the authors used Naive Byes and TAN [Tree Augmented Naive] classifiers [32]. In [33] [34], classification is performed using SVM (Support Vector Machine) classifier which yielded better results. In [35], multi-stage binary code (MSBP) is proposed and experimented on two different kinds of approaches, holistic and division based which shows 96% and 60% accuracies respectively.

## 3 Proposed Methodology

The major components of our system are discussed in following subsections.

### 3.1 Preprocessing

Pre-processing step includes face detection on the images, applying the elliptic mask, and then cropping the desired facial area. There are many methods for detecting the face in the literature, for example, Memat et al. [36] detected face edges employing high pass filter of the wavelet transform and SVM was used for classification purpose. We have used Viola-Jones face detector method [37] because of its robustness on all kinds of images. Elliptic mask [38] was applied for highlighting the desired oval faced area and then it was cropped. The extracted face and cropped elliptic region are shown in Figs. 1(a) and (b) respectively. Where, Figs. 1(c) and (d) show results after applying Equation 1 in Figs. 1(a) and (b) respectively. Images are also properly resized basis on their aspect ratios [39].

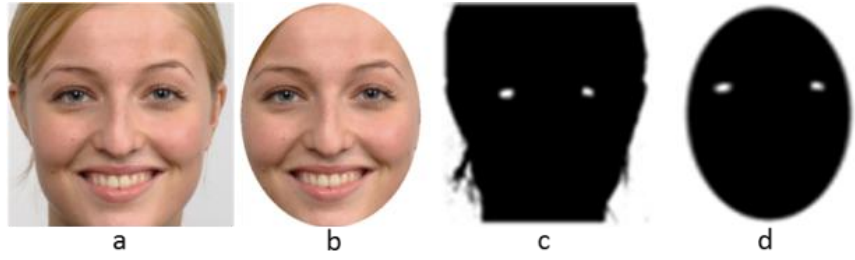

**Fig. 1** (a) Extracted face (b) Elliptic mask applied (c), (d) Images b and d after applying Equation 1.

*3.1.1 Eye Detection*

Turkan et al. [40][41] presented multiple studies for eye detection based on the high-pass filter of a wavelet transform. Eyes were also tracked using Markov model by Bagci et al. [42]. Moreover, for eye detection, Nasiri et al. [43] proposed the idea that eye region has low Cr and high Cb values. They performed EyeMapC and EyeMapL processes (explained in their article [44]) and then they concatenated both. This involves lengthy morphological operations, extra phases, different thresholding and geometrical testing. In an attempt to overcome these extra operations we got an anomaly between our methodology and the one mentioned by Nasiri et al. In contrast with their approach, employing Equation 1, we increased the value of Cr and suppressed the value of Cb from the YCbCr color domain [45] images. The resulting images are shown in Figs. 1(c) and 1(d). We named these methods as ECrR and SCrB as mentioned in section 1. We got equivalent accuracies (~98%) in results with fewer operations.

$$Eyemap = (Cr(x,y))^2 \left((Cr(x,y))^2 - (Cr(x,y)/Cb(x,y))\right)^4 \qquad (1)$$

For locating the left eye, the image was scanned from left to right, searching for white pixels. The process was repeated for the right eye by scanning from right to left. Then based on the eyes location, the location of eyebrows was searched out.

*3.1.2. Cropping Facial Region*

For eyebrows region cropping, the eyes location was calculated as subsection 3.1.1. In the end, we obtain the images depicted in Figs. 5 and 6. The Wrinkles' region is also cropped out based on eyes locations. For lips region extraction, the oval-shaped face area is resized to 381X281 pixels for scaling all the images due to this scaling we cropped the lips area for all the images exactly from the same location where it always located. Then, for obtaining the location of the lips area only, we cropped the area from the row pixels 250 to 381 and column pixels from 1 to 281 because the images are properly scaled and always lips are located in the same area. This gives us the desired location of the lips as represented in Figs. 2(a) and 3(a).

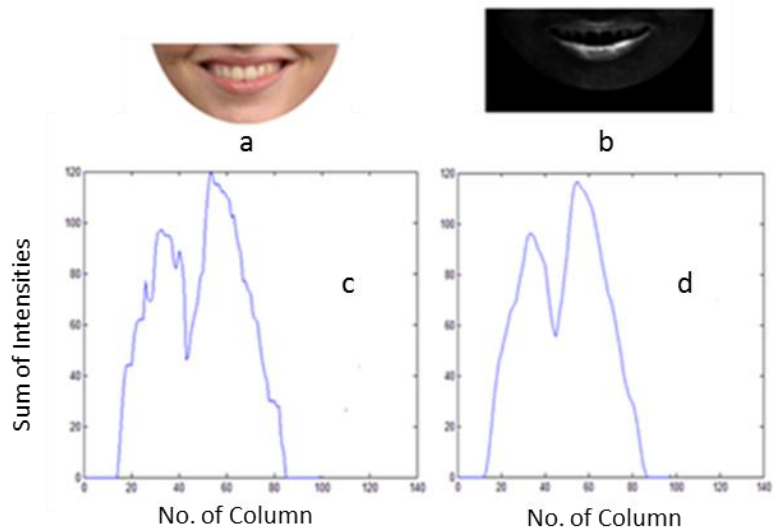

**Fig. 2** Representing the steps of lips feature extraction in case of happy emotion (a) Cropped lips area (b) Lips feature extracted after applying eq.2 and eq.3 (c) Represents intensities of white pixels (d) Smoothing Applied

*3.2 Feature Extraction*

As mentioned in Section 1, the emotions are identified based on mouth opening (MO), eyebrows constriction (EBC), Eyebrows mean (EBM), Lips corners (LC) and forehead wrinkles (W). The extraction of these features is explained below:

*3.2.1. Mouth Opening*

Lips opening and closing plays a decisive role in emotion expression so we have considered this as a salient feature for emotion detection. As mentioned by Aleksandra et al., the results using the RGB system are not as accurate as YCbCr color space especially in case of skin detection [44][45]. Since lips have relatively more redness value as compared to other facial regions; therefore, further processing is performed on the Cr component of the image [45]. Cr values are highlighted using following Equation:

$$Mouthmap = (Cr(x,y))^2 \left( (Cr(x,y))^2 - \frac{nCr(x,y)}{Cb(x,y)} \right)^2 \tag{2}$$

The value of n is computed as

$$n = 0.95 \left( \frac{(1/k) \sum Cr(x,y)^2}{(1/k) \Sigma \left( Cr(x,y) / Cb(x,y) \right)} \right) \tag{3}$$

Where, k is the total number of pixels in an image. This technique is applied only to the lower half face area to reduce the computational cost. The outputs after highlighting the Cr values are shown in Figs. 2(b) and 3(b). In the next step, Sobel edge detector was applied in Figs. 3(b) to extract lip edges. To reduce noise, we have applied morphological opening. Sum of pixels intensities of each row is computed and then plotted as in Figs. 2(c) and 3(c). Smoothing was applied on Figs. 2(c) and 3 (c) to eliminate small peaks.

The results of smoothing are shown in Figs. 2(d) and 3(d). The number of peaks is calculated by finding local maxima which are used to decide whether the mouth is opened or closed. In case of opened mouth, we get two peaks as depicted in Figs. 2(c) and (d). In case of closed mouth, we get only one peak after applying smoothing as depicted in Figs. 3(c) and 3(d). In some cases, we get two peaks, representing open mouth for disgust or surprised etc. Therefore, we have to consider the degree of mouth opening too to classify among these cases. The degree of mouth opening can simply be measured by the distance between the two peaks. Therefore, mouth opening alone cannot be used to classify emotion and we have to consider the degree of mouth opening to classify the emotion. The degree of mouth opening can simply be measured by the distance between the peaks and in case of only one peak, it is considered mouth closed.

In Figs. 2(c), 2(d), 3(c) and 3(d), the x-axis represents a number of column and y-axis shows the sum of intensities of white pixels in that column corresponding to the processed images.

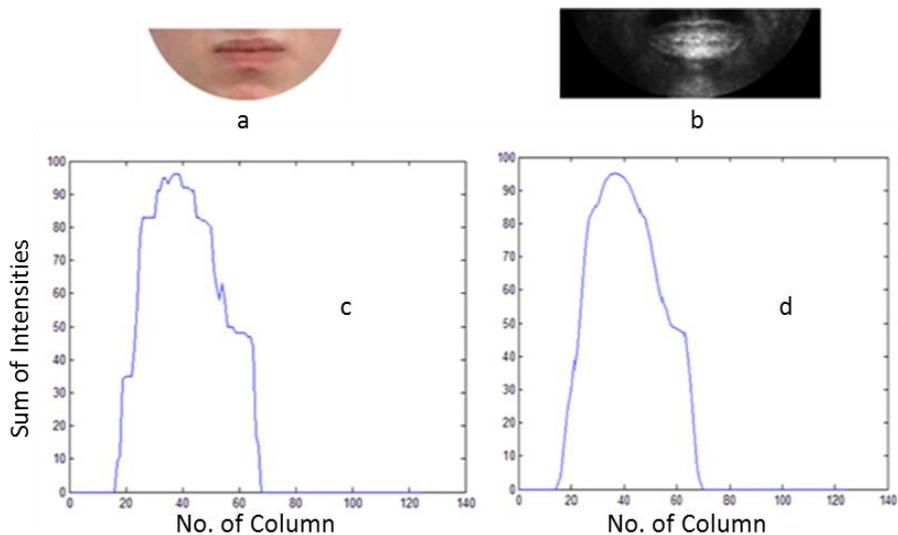

**Fig. 3** Representing the steps of lips features extraction (a) Cropped lips area of disgust emotion (b) Lips feature extracted after applying eq.2 and eq.3 (c) Represents intensities of white pixels (d) Smoothing Applied

*3.2.2. Eyebrows Constriction*

We have proposed two methods for eyebrows feature extraction, the methods are given below:

➤ **Method 1:**

Different facial areas were cropped including eyebrows region as explained in 3.1.2. The cropped regions were converted into greyscale and morphological gradient operation [39] was applied, specifically, to highlight the eyebrow area as shown Figs. 4(c).

Subsequently, the image intensities values were scanned column wise and the mean position of high-intensity values of each column was calculated where high intensities refer to that area where the eyebrows pixels are present in the image. Considering the mean location as a central point a line was plotted. The process was repeated for all columns which ultimately gives us the average position of the eyebrow.

We have to normalize this value according to the size of the cropped region to avoid any error due to regions of different size. This normalization can simply be done by dividing the mean position of the eyebrow by the height of the window represented in Figs. 4.

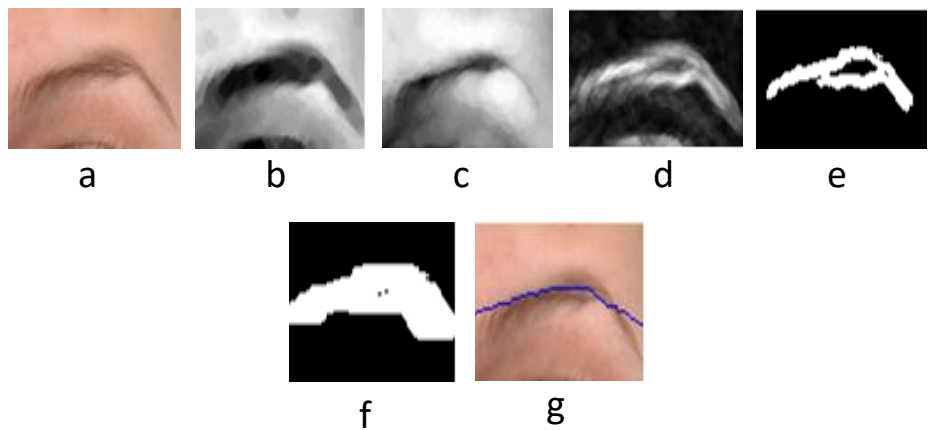

a    b    c    d    e

f    g

**Fig. 4** Representing the steps of eyebrows features extraction. Applied Method 1, a) Cropped eyebrow, b) Grayscale image, c), d), e) and f) Applied morphological gradient, g) Plotted line, on the calculated mean positions of high intensities.

➢ **Method 2:**

Another technique, DCL for eyebrows feature extraction was also proposed that gives better results. First of all the RGB image was converted into a binary with a threshold 0.5, as in Figs. 5 (a) and (b). The Sobel edge detector was applied to the binary image and the noise was removed by using the morphological opening methodology. Figs. 5(c) shows the applied Sobel edge detection and removed noise. The first white pixels were searched column wise and a line was plotted as shown in Figs. 5(d). Afterward, the degree of curvature of the line was calculated by taking the derivative of the plotted line on each point and adding the absolute values of the derivative at every point. Then we have computed the sum of all the calculated values. Equation 4 represents the procedure.

$$C = \left| \left( \Delta \left( central\ line \right) \right) \right| \qquad (4)$$

equation 4: should be L1 norm. Then the degree of curvature of the eyebrow was normalized by dividing it by the length of the line representing the eyebrow as in Equation 5:

$$N = \sum C\ /\ length\ (C) \qquad (5)$$

By dividing the sum by the length we get normalized curvature of the eyebrow. This normalization is performed to avoid any error in this feature due to different lengths of the eyebrows.

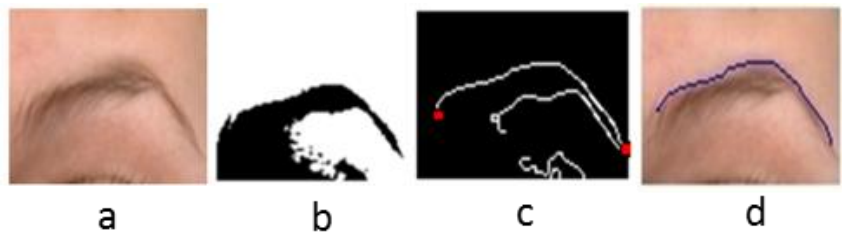

a  b  c  d

**Fig. 5** Representing the steps of eyebrows features extraction applying method 2. (a) Cropped eyebrow (b) Converted into a binary image (c) Applied Sobel edge and morphological opening on binary image (d) Searched first white pixels column wise and plotted a line.

*3.2.3 Mouth Corners Detection*

The lips region was cropped to calculate the distance between mouth corners to be used as a feature. For this feature extraction, the cropped image is initially converted into grayscale and then Equation 6 is applied that highlights the lips corner only and suppresses the other parts of the image. In the first step, we performed dilation of the lips area using spherical structuring element in the binary mouth image due to which mouth corners are covered. As mouth corners are the darkest points in the dilated area. Therefore, we can make such a map which highlights the darkest points using equation 6.

$$CornersArea = (255 - grayLips(x, y))^6 \qquad (6)$$

Where, graylips(x,y) is the Luminance part of the YCbCr image that is present in the (x,y) coordinates of the image. After highlighting the corners, the global highest value from the lip region image was searched. While scanning the image from left to right, the point where the value is less than one half of the global highest value is the required left corner of the image. Similarly, the image is scanned from right to left and the required right corner of the mouth is found. Subsequently, the distance between these two corners is calculated. This distance serves as a feature to classify the emotion. The results after the application of the above procedure can be seen in Figs. 6(a) and (b).

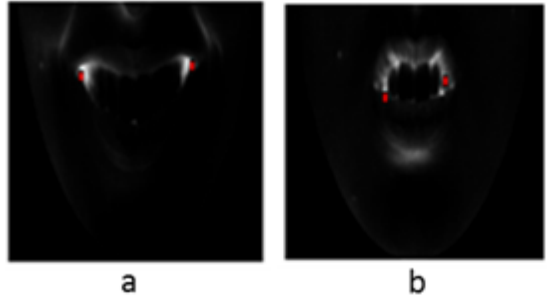

**Fig. 6** Representing the steps of mouth corner detection (a) detected corners in case of happy emotion, (b) detected corner in case of disgust emotion.

*3.2.4 Forehead Wrinkles*

Different methods were applied to detect the intensity of wrinkles like Fourier transform [39], Canny edge detection [39], and point detection [39] but the best results were achieved using Canny edge detector. While observing different images, it is concluded that higher intensity of forehead wrinkles is present in the forehead area located between two eyebrows. Therefore in our experimentation; we have taken the said areas into account for wrinkles detection. For example Figs. 7(a) shows face with wrinkles between the eyebrows and Figs. 7(b) shows face

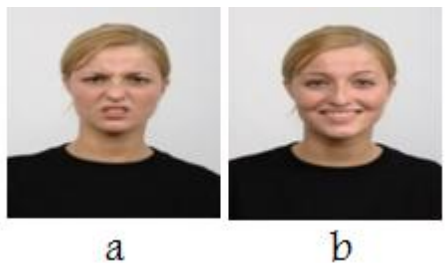

with no wrinkles.

**Fig.7** Representing (a) presence and (b) absence of wrinkles between eyebrows.

As eyes location is already known as discussed in section 3.1.1, the area containing wrinkles was cropped. The Figs. 8(a) and (b) represent the cropped area of wrinkles. The Figs. 8(c) and (d) illustrate results after applying Canny edge detector. The intensity of wrinkles can then be calculated by simply adding the white pixels in the images obtained after applying Canny edge detector.

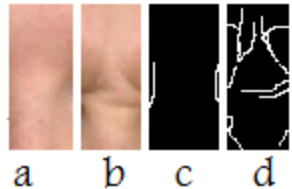

**Fig. 8** Representing the intensity of wrinkles(a) Absence of wrinkles in case of happy emotion (b) Presence of wrinkles in case of disgust emotion (c), (d) Canny edge detection applied on (a) and (b) respectively

## 4   Databases

I.  The Radboud Faces Database (RaFD) (Langer et al. [11]) is a set of pictures of 67 models (including Caucasian males and females, Caucasian children, both boys and girls, and Moroccan Dutch males) displaying eight emotional expressions. The RaFD is a high-quality faces database, which contains pictures of eight emotional expressions. Accordingly, to the Facial Action Coding System, each model was trained to show the following expressions: Anger, disgust, fear, happiness, sadness, surprise, contempt, and neutral. Each emotion was shown with three different gaze directions and all pictures were taken from five camera angles simultaneously.

II. The database Static Facial Expressions in the Wild (SFEW) [12] has been developed by selecting frames from Acted Facial Expressions in the Wild (AFEW) [13] which is a dynamic temporal facial expression data corpus consisting of close to real-world environment extracted from movies. The database covers unconstrained facial expressions, varied head poses, large age range, occlusions, varied focus, and different resolution of the face and close to real-world illumination. Frames were extracted from AFEW sequences and labeled based on the label of the sequence. In total, SFEW contains 700 images and that has been labeled for six basic expressions angry, disgust, fear, happy, sad, surprise and the neutral class and was labeled by two independent labelers.

III. The Japanese Female Facial Expression (JAFFE) database [14] contains 213 images of 7 facial expressions (6 basic facial expressions + 1 neutral) posed by 10 Japanese female models. Each image has been rated on 6 emotion adjectives by 60 Japanese subjects.

## 5 Classification

The classification has been performed using individual classifiers as well as by MV and WMV. We have defined some rules for classification by dividing the images of the databases. For training and testing, we have chosen 600 images randomly for training purpose from the databases and defined the rules and rest of the images are used as a testing images. Classifier combination has a broad range of applications including recognition of face, pattern, fingerprints, character, speech and medical images. The patterns classified by individual classifier may be accurate in a certain uncontrolled environment or partially occluded images and all of the individual classifiers is good enough to perform all the tasks with adequate accuracy in an uncontrolled environment. Additionally, for more accuracy, different classifiers need to be combined to achieve reasonable recognition rates. The MV and WMV rules were chosen for classification purpose based on ensemble classifier [9] [10] as well as Human Emotion Sensitivity (HES).

After analyzing all images in the databases, the empirical rules were finalized for classification which are based on HES system for classification, such as, IF (Mouth Open) & (Lips Corners

Wide) & (No Wrinkles) & (No Curve in Eyebrows) & (Eyebrows stretched), THEN considered happy, Else disgust. Then it will check the rules for all remaining emotions. Rules for happy and disgust are mentioned below and rest are given in Table 1.

- If the value of eyebrow Curvature is less than 0.5 then emotion is classified as disgust emotion, else happy. The value of Curvature is calculated using Equations 4 and 5.
- If the mean of a line is greater than 0.7 then happy else it is considered as disgust. The mean of a line for eyebrow is calculated by method 2 in Section 3.2.2.
- If the intensity of wrinkles is greater than 200 pixels then disgust emotion else happy. Wrinkles intensities are calculated in Section 3.2.4.
- If mouth opening is less than 25 pixels then the emotion is classified as disgust emotion else happy. The value of mouth opening is calculated using Equations 2 and 3.
- If the distance between lips corners is greater than 50, then emotion is classified as happy else it is considered as a disgust emotion. And the value of the lips corners is calculated using Equation 6.

**Table 1** Representing rules for emotions.

| MO | LC | W | EBC | EBM | Classified emotion |
|---|---|---|---|---|---|
| <25 | <50 | >200 | < 0.5 | < 0.7 | Disgust |
| >25 | >50 | > 200 | >0.5 | >0.7 | Surprise |
| <25 | <50 | <200 | <0.5 | <0.7 | Angry |
| <25 | <50 | <200 | >0.5 | >0.7 | Neutral |
| >25 | >50 | <200 | > 0.5 | >0.7 | Happy |

Our initial experiments were to classify emotion against individual feature i.e. to know how well an individual feature is useful for classifying a specific emotion. On later stages, we considered the decision of individual classifiers using two techniques, namely MV and WMV to classify an emotion based on all the features like MO, EBC, EBM, LC, and W collectively. First, we have evaluated the system based on its accuracy rate of feature detection, and then on the feature classification. In the end, the results are compared with the previous methodologies.

For feature extraction, we applied Equation 1 on Figs. 1(a) and 1(b), the overall feature detection accuracy achieved was 98%. While applying Equations 2 and 3 on Figs. 2(a) and 3(a) yielded 95% overall feature detection accuracy for lips. Whereas, the feature detection accuracy rate for corners detection after applying the Equation 6 on Figs. 2(a) and 3(a) is 98.9%. For eyebrows detection using method 1 and method 2, the accuracy rates are 85% and 94% respectively. The summary of results is represented in Table 2.

**Table 2** Representing feature Precision rates.

| Methods | Precision rates (%) |
|---|---|
| Eye detection | **98** |
| Lips detection | **95** |
| Corners detection | **98.9** |
| Eyebrow detection(1) | **85** |
| Eyebrow detection(2) | **94** |

When an image is given to our system, the system generates feature values mentioned in the rules above which can be called belief values. Initially, we have presented the results of classification by using individual features and their respective rules as classifiers to categorize the emotion. The results of such individual classifiers are shown in table 2. Which are computed as:

$$A = \sum (All\ positive\ response\ /\ total\ number\ of\ images) * 100 \qquad (7)$$

Where, A represents the accuracy rate against a specific feature. As evident from table 2, the results against W is not that good as compared to those of LC and MO. This drop in recognition rates is due to the inherent error in detection of wrinkles, which is a consequence of inaccurate cropping of that region. Since the wrinkles region was cropped automatically based on the eye locations and the distance between them varies from person to person, therefore, it causes an error in feature detection. If the wrinkles region is cropped manually, the recognition rates against this feature alone give around 90% accuracy. As per results are shown in table 2, wrinkles alone does not give good recognition rates.

Table 3 Accuracy rates of individual features Happy(H), Disgust(D), Surprise(S), Angry(A) and Neutral(N).

| Features | EBM (%) | LC (%) | EBC (%) | W (%) | MO (%) |
|---|---|---|---|---|---|
| Accuracy rates(D) | 70 | 99 | 99 | 61 | 99 |
| Accuracy rates(H) | 75 | 98 | 90 | 67.46 | 97 |
| Accuracy rates(S) | 80 | 98 | 98 | 63 | 96 |
| Accuracy rates(A) | 80 | 97 | 98 | 65 | 98 |
| Accuracy rates(N) | 90 | 99 | 99 | 70 | 97 |

In case of individual classification, the accuracy rates of happy, disgust, surprised, angry, and neutral are calculated and mentioned in table 3, for EBM, LC, EBC, W, and MO.

Table 4 Accuracy rates of individual features and their computed weights for WMV.

| Features | EBM | LC | EBC | W | MO |
|---|---|---|---|---|---|
| Accuracy rates(D) | 70 | 99 | 99 | 61 | 99 |
| Weights(WMV,D) | 0.183 | 0.236 | 0.236 | 0.160 | 0.236 |
| Accuracy rates(H) | 75 | 98 | 90 | 67.46 | 97 |
| Weights(WMV,H) | 0.1943 | 0.2305 | 0.1787 | 0.1735 | 0.222 |

We have analyzed the performance of our classifiers by using different combinations of individual classifiers using both MV and WMV. These classifiers were tested against all emotion. For WMV, the empirically calculated weights for each of the individual classifiers against all the emotion are given in table 4. The weights of individual classifiers are calculated through Equation 8.

$$Weight\ of\ the\ feature = \frac{Accuracy\ rate\ of\ the\ feature}{Sum\ of\ accuracy\ rates\ of\ all\ features} \quad (8)$$

**Table 5** Accuracy rates of individual features and their computed weights for WMV.

| Features | EBM | LC | EBC | W | MO |
|---|---|---|---|---|---|
| Accuracy rates(D) | 70 | 99 | 99 | 61 | 99 |
| Weights(WMV,D) | 0.183 | 0.236 | 0.236 | 0.160 | 0.236 |
| Accuracy rates(H) | 75 | 98 | 90 | 67.46 | 97 |
| Weights(WMV,H) | 0.1943 | 0.2305 | 0.1787 | 0.1735 | 0.222 |

In one of the combinations, the most reliable features namely lip corners (LC) and mouth opening (MO) were combined whose accuracy rate is 99% against disgust emotions and 98.5% against happy emotions. Another combination is formed by merging one weak feature with two strong feature which is eyebrows mean (EBM), eyebrows curvature (EBC) and MO whose accuracy rates are 97% against disgust emotions and 98% against happy emotions. In two other combinations, two strong and one weak or strong feature were merged, which are LC and MO with either EBM or EBC whose accuracy rates are 99% and 97% or 99.5% and 98.4% respectively for the disgust and happy emotions. In another combination, the weakest classifier merged with two strong classifiers which are MO, LC and wrinkles (W) whose accuracy rate surprisingly was the best on average for all emotions. A detailed summary is given in table 6 and calculated average emotion classification accuracy for all emotions is 94**%** which is very promising.

**Table 6** Accuracy rates with different combinations using WMV.

| Features | MO+LC | EBC+EBM+MO | LC+MO+EBM | LC+MO+EBC | MO+LC+W | EBC+W | EBC+EBM | Sum | Average |
|---|---|---|---|---|---|---|---|---|---|
| Accuracy rates (D) | 99 | 97 | 99 | 99.5 | 98 | 75 | 99 | 666.5 | 95.21429 |
| Accuracy rates (H) | 98.5 | 98 | 97 | 98.4 | 97.9 | 69 | 96 | 654.8 | 93.54286 |
| Accuracy rates (S) | 98 | 97 | 98 | 97.7 | 98 | 70 | 97 | 655.7 | 93.67143 |
| Accuracy rates (A) | 96 | 99 | 98 | 97 | 99.5 | 71 | 96 | 656.5 | 93.78571 |
| Accuracy rates (N) | 98 | 97 | 99 | 97 | 99 | 75 | 98 | 663 | 94.71429 |
| Sum: | | | | | | | | | 470.9286 |
| Average: | | | | | | | | | 94.18571 |

Two more combinations consisting of one weak and one strong classifier each were also tested, which are EBC along with wrinkles (W) and EBC along with EBM.

The other ensemble approach which was used in this study is the majority voting (MV) approach. This is emotion based on the decision of the majority of the individual classifiers. As we have five individual classifiers so the emotion classified by a minimum of three classifiers will be considered the decision of MV classifier.

## 4 Results and Comparisons

As illustrated in table 6, it can be noticed that our accuracy rate lies always above 80%. Arshid et al. [35] recently used different techniques including local bit pattern. The average accuracy results for the Local bit pattern using Naive Bayes are 65% which are depicted in table 1 of ref. [35]. S. Zhang et al. [45] recently demonstrated 90% accuracy rates using hidden Markov models (HMM). Uddin et al. [46] recently reported different approaches for facial expression recognition. Their results are summarized in tables 1–5 in ref. [46]. These results are based on Principal Component Analysis (PCA) with Hidden Markov Model (HMM), Principal Component Analysis and Local Directional Pattern (PCA-LDA) with HMM, Independent Component Analysis (ICA) with HMM, LBP with HMM, and LDP with HMM with accuracy rates of 58%, 61%, 80%, 81% and 82% respectively.

In our proposed methodology through the individual classifier, MV and WMV, the average accuracy rates are 89%, 93%, and 94%. The results comparison is given below we have chosen the same database from the literature but as in the literature, no one has used this kind of classifier for emotion recognition purpose that's why I compared the results with different classifiers.

Table 7 Comparision of the proposed methods with other state of the art techniques.

| Ref. | Database | Feature Extraction | Classifer | Avg.Recgnition Rate(%) |
|---|---|---|---|---|
| [35] | SFEW | LBP | Naïve Bayes | 65 |
| [35] | SFEW | LBP | Bagging | 69 |
| [45] | JAFFE | LBP+LFDA | SVM+HMM | 90 |
| [46] | SFEW | LBP | HMM | 81 |
| [47] | JAFFE+SEEW | LBP | AdaBoost | 92 |
| [48] | JAFFE | DCT | SVM+HMM | 61 |
| [49] | JAFFE | Gabor+PCA | SVM | 82 |
| Ours | JAFFE +SFEW | Introduced Method. | WMV | 94 |
| Ours | JAFFE+SFEW | Introduced Method. | MV | 92 |
| Ours | RAFD | Introduced Method. | MV | 94 |
| Ours | RAFD+JAFFE+SFEW | Introduced algo. | Individual Feature | 89 |

## 5 Conclusion

We have proposed a solution to the problem of automated emotion recognition by introducing different methods. Our emotion detection system is a reliable system which gives accuracy rate around 94%. Our proposed technique is best for partially occluded images as our proposed system also can identify the emotion based on the individual features. The overall accuracy rates are found to be 94% which shows immense potential.

In our opinion, for making capacity for performance improvement, we can induce these methods in Convolutional Neural Networks (CNN) for training purpose which can provide the best results on the testing images for facial emotion discrimination which can improve the performance.

## 6 Acknowledgments

Higher Education Commission have fully funded for the research work completed.

1388

[12]. Dhall A, Goecke R, Lucey S and Gedeon T 2011 Static Facial Expression Analysis In Tough Conditions: Data, Evaluation Protocol and Benchmark IEEE International Conference on Computer Vision, Workshops (ICCV Workshops), 2106-2112

[13]. Dhall A, Goecke R, Lucey S, and Gedeon T 2011 Acted Facial Expressions in the Wild Database In Technical Report TR-CS-11Australian National University, Canberra, Australia, 2, 1

[14]. Lyons, M., et al. 1998 Coding facial expressions with gabor wavelets. In: Automatic Face and Gesture Recognition IEEE Proceedings on Third IEEE International Conference

[15]. Ekman P and Friesen W 1978 Facial Action Coding System: A Technique for the Measurement of Facial Movement  Consulting Psychologists Press, Palo Alto

[16]. Ekman P 1993 Facial expression and emotion American Psychologist, 48(4), 384

[17]. Lyons M J, Budynek J, and Akamatsu S 1999 Automatic classification of single facial images IEEE Transactions on Pattern Analysis and Machine Intelligence, 21(12), 1357-1362

[18]. Kanade T, Cohn J F, and Tian Y 2000 Comprehensive database for facial expression analysis Proc. Fourth IEEE International Conference on Automatic Face and Gesture Recognition, 46-53

[19]. Fasel B and Luettin J 2003 Automatic facial expression analysis: a survey," Pattern Recognition, 36(1), 259-275

[20].  Hageloh F, Sande K V D, and  Valenti R 2005 Automatic facial emotion recognition Universiteit van Amsterdam computing Surveys (CSUR), 35(4), 399-458

[21]. Alvino C, Kohler C, Barrett F, Gur E R, Gur C R, Verma R 2007 Computerized measurement of facial expression of emotions in schizophrenia Journal of neuroscience methods, 163, (2), pp. 350–361

[22]. Bashyal S, Venayagamoorthy K G 2008 Recognition of facial expressions using Gabor wavelets and learning vector quantization Engineering Applications of Artificial Intelligence, 21, (7), pp. 1056–1064

[23]. Wong J J, Cho S Y 2009 A local experts organization model with application to face emotion
24